\documentclass[letterpaper, 10 pt, conference]{ieeeconf}  

\IEEEoverridecommandlockouts                              

\usepackage{cite}
\usepackage{amsmath,amssymb,amsfonts}
\usepackage{algorithmic}
\usepackage{subcaption}
\usepackage{graphicx}
\usepackage{textcomp}
\usepackage{xcolor}
\usepackage{booktabs,floatrow}
\usepackage{comment}
\usepackage{amsmath}
\usepackage{url}
\usepackage{makecell}
\usepackage{svg}
\usepackage{multicol}
\usepackage{adjustbox}
\usepackage{epstopdf}
\usepackage{comment}
\usepackage{hyperref}
\hypersetup{
    colorlinks=true,
    linkcolor=blue,
    }

\title{\LARGE \bf
WiDEVIEW: An UltraWideBand and Vision Dataset for Deciphering Pedestrian-Vehicle Interactions
}

\author{Jia Huang$^{1}$, Alvika Gautam$^{2}$, Junghun Choi$^{3}$, and Srikanth Saripalli$^{4}$
\thanks{*This work is funded by Hyundai Motor Company}
\thanks{$^{1}$Jia Huang is with the J. Mike Walker ’66 Department of Mechanical Engineering, Texas A\&M University, USA
        {\tt\small jia.huang@tamu.edu}}%
\thanks{$^{2}$Alvika Gautam is with the J. Mike Walker ’66 Department of Mechanical Engineering, Texas A\&M University, USA
        {\tt\small alvikag@tamu.edu}}%
\thanks{$^{3}$Junghun Choi is with Hyundai Motor Company, Korea
        {\tt\small whvlfwo@hyundai.com}}%
\thanks{$^{4}$Srikanth Saripalli is the faculty with the J. Mike Walker ’66 Department of Mechanical Engineering, Texas A\&M University, USA
        {\tt\small ssaripalli@tamu.edu}}%
}

\begin{document}

\maketitle
\thispagestyle{empty}
\pagestyle{empty}

\begin{abstract}
Robust and accurate tracking and localization of road users like pedestrians and cyclists is crucial to ensure safe and effective navigation of Autonomous Vehicles (AVs), particularly so in urban driving scenarios with complex vehicle-pedestrian interactions. Existing datasets that are useful to investigate vehicle-pedestrian interactions are mostly image-centric and thus vulnerable to vision failures. 
In this paper, we investigate Ultra-wideband (UWB) as an additional modality for road users' localization to enable a better understanding of vehicle-pedestrian interactions. We present WiDEVIEW\footnotemark[1], the first multimodal dataset that integrates LiDAR, three RGB cameras, GPS/IMU, and UWB sensors for capturing vehicle-pedestrian interactions in an urban autonomous driving scenario. Ground truth image annotations are provided in the form of 2D bounding boxes and the dataset is evaluated on standard 2D object detection and tracking algorithms. The feasibility of UWB is evaluated for typical traffic scenarios in both line-of-sight and non-line-of-sight conditions using LiDAR as ground truth. We establish that UWB range data has comparable accuracy with LiDAR with an error of 0.19 meters and reliable anchor-tag range data for up to 40 meters in line-of-sight conditions. UWB performance for non-line-of-sight conditions is subjective to the nature of the obstruction (trees vs. buildings). Further, we provide a qualitative analysis of UWB performance for scenarios susceptible to intermittent vision failures. 
\end{abstract}

\section{Introduction}
Intelligent and autonomous driving systems have grown rapidly over the past decade resulting in highly capable autonomous vehicle systems performing perception, planning and control tasks in a sophisticated manner. Yet these systems are faced with safety concerns especially in complex urban driving scenarios. In such scenarios, vehicle interaction with other road users is a critical aspect that must be taken into consideration to ensure safe navigation. Safe interaction includes vehicle being able to detect, track and localize vulnerable road users such as pedestrians and cyclists, recognize or anticipate their actions and make appropriate decisions to avoid collisions.

Comprehensive investigation of pedestrian-AV interactions is challenging as real-world studies get typically limited due to ethical, control and safety reasons. A significant body of literatures has focused on understanding these interactions and pedestrians'/cyclists' tracking and localization through numerous large scale datasets (single as well as multimodal sensor data). 
For urban driving scenarios, most existing datasets are image-centric \cite{JAAD_cite1,JAAD_cite2,psi_dataset,STIP} while only limited ones provide additional sensor data for depth information such as LiDAR in \cite{pedx} and on-board diagnostics (OBD) data in \cite{PIE_dataset} for vehicle information like GPS coordinates and speed. Image-centric single modality sensing configurations are fragile and have limitations in adverse lighting conditions, blind spots, etc. and are thus prone to intermittent failures. 
Although there has been an increasing number of multimodal urban autonomous driving datasets in the past few years \cite{kitti,nuscenes,waymo}, they are typically not intended for studying vehicle-pedestrian interactions but tend to focus on general tasks like object detection and tracking, semantic segmentation for scene understanding, etc. 

In this paper, we investigate the use of Ultra-wideband (UWB) sensing as an additional modality for road users' information. We hypothesise that augmenting UWB can provide improvements where vision is susceptible to intermittent detection/tracking failures in scenarios like bad lighting conditions, obstructions, vehicle blind spots, intersections etc by providing continuous and in some cases more timely information about road users. Further, we release the first benchmark dataset WiDEVIEW\footnote{\url{https://github.com/unmannedlab/UWB_Dataset}} which includes LiDAR, three RGB cameras, high precision Global Positioning and Inertial Navigation System (GPS/INS), and UWB sensors and focuses on vehicle-pedestrian interactions. 
In addition to raw sensor data, the ground truth annotations are provided for RGB camera images for the first release, more annotations will be added continuously. 
 

The contribution of our work can be summed up as follows:
\begin{itemize}
    \item To the best of authors' knowledge, this is the first multimodal dataset that integrate LiDAR, three RGB cameras, GPS/IMU and UWB with a focus on vehicle-pedestrian interaction in urban/campus autonoumous driving scenarios. 
    \item The dataset consists of 21 sequences of sensor data captured in Robot Operating System (ROS) bag format while driving on campus and busy road. 2D bounding boxes tracking annotations are provided for every fifth frame of the front middle camera.
    \item Benchmark analysis for 2D object tracking and UWB Tag (road users) localization are established by utilizing the standard object detection/tracking algorithms and the trilateration algorithm, respectively.
    \item Qualitative evaluations are carried out to compare vision and UWB sensing in typical vehicle-pedestrian interactions scenarios.
\end{itemize}
The aim of this dataset is to provide extensive scenarios capturing vehicle-pedestrian interactions in urban driving setting, which can aid in developing better pedestrian tracking and localization, and eventually safe autonomous navigation algorithms by augmenting UWB and using the complimentary properties of UWB sensing with vision and LiDAR data.
\section{Related Work}
\subsection{Related Datasets}
Urban environment datasets for autonomous driving include single as well as multiple sensor modalities. Single modality ones \cite{camvid,vistas,BDD100K} include camera data only. Examples of multimodal datasets which contain image data, range sensor data (LiDAR, RADAR) include \cite{kitti}, \cite{KAIST},\cite{apolloscape},\cite{h3d} as well as large scale ones like nuScenes\cite{nuscenes}, Waymo Open\cite{waymo}, Lyft L5 \cite{Lyft}. Overall, these datasets focus more on scene understanding research for autonomous vehicles.
 
From the perspective of vehicle-pedestrian interaction, JAAD \cite{JAAD_cite1} is one of the initial datasets to study pedestrian and driver behaviors at the point of crossing. Another large scale dataset PIE \cite{PIE_dataset} by Rasouli et al. also includes pedestrian crossing intention and vehicle information from OBD sensor. STIP \cite{STIP} uses three front RGB cameras and contains pedestrian bounding boxes and labels of crossing/not-crossing the street. TITAN\cite{2020titan} uses camera with embedded IMU sensor which records synchronized odometry data for ego-motion estimation and provides 50 labels including vehicle states and actions, pedestrian age groups and targeted pedestrian action attribute. PedX\cite{pedx} is a large-scale multimodal dataset of pedestrians at complex urban intersections captured using two pairs of stereo cameras and four Velodyne LiDAR sensors along with 2D and 3D labels of pedestrians. PSI \cite{psi_dataset} contains scene videos, GPS coordinates and vehicle speed as well as two innovative labels other than common computer vision annotations. A summary of datasets along with WiDEVIEW with the corresponding sensing modalities and annotations is provided in Table \ref{tab:dataset_table}.

\begin{table}[!ht]
\noindent
\scriptsize
\resizebox{\columnwidth}{!}{
\begin{tabular}{cccccc} \hline
        %
         Dataset & Sensors & Annotations  & No. of Classes & Year\\
         \hline
         JAAD\cite{JAAD_cite1} & Camera & Bbox  & 3 & 2017  \\
         PIE\cite{PIE_dataset} & Camera, OBD & Bbox & 6 & 2019 \\
         PedX\cite{pedx}  &Camera,LiDAR  & Pose, Seg & 1 & 2019 \\
         TITAN\cite{2020titan} & Camera,IMU &Bbox  & 3 & 2020 \\
         STIP\cite{STIP} & Camera & Bbox  & 1 & 2020 \\
         PSI\cite{psi_dataset} & Camera & Bbox, Pose, Seg & 20 & 2021 \\
         \hline
         KITTI\cite{kitti}& Camera,LiDAR& Bbox&  8& 2020\\
         KAIST\cite{KAIST}& Camera,LiDAR& Bbox&  5& 2020\\
         ApolloScape\cite{apolloscape}& Camera,LiDAR &Bbox, Seg& 8-35& 2020\\
         H3D\cite{h3d}& Camera,LiDAR& Bbox& 8 & 2020\\
         NuScenes\cite{nuscenes} & Camera,LiDAR,Radar& Bbox& 23& 2020 \\
         OpenMPD\cite{openMPD} & Camera,LiDAR& Bbox, Segmentation& 6/11& 2020 \\
         Waymo\cite{waymo}& Camera, LiDAR& Bbox & 4& 2020\\
         Lyft L5\cite{Lyft}& Camera,LiDAR& Bbox & 9& 2020\\
         \hline
         \textcolor{blue}{WiDEVIEW(ours)} & Camera, LiDAR, UWB & Bbox & 1 &  2023 \\
        \hline
\end{tabular}}
\caption{Comparison of popular autonomous vehicle research datasets. The top half are vehicle-pedestrian datasets. The bottom half are major multimodal datasets that include but do not focus on vehicle-pedestrian interactions.}
\label{tab:dataset_table}
\end{table}

\subsection{UWB based Localization}
UWB ranging has been used in both ground and aerial systems, often when fused with data from other sensors, for accurate localization especially in GNSS unreliable or denied environments \cite{sun2019ins,ghanem2020testing,nguyen2019integrated} and collaboration in multi-robot systems \cite{shule2020uwb,jacob_collaborative}. There are only a limited number of open datasets that focus on UWB aided localization and most of them are limited to either indoor applications or aerial vehicle use cases \cite{raza2019dataset,queralta2020uwb,nguyen2022ntu}. To the best of our knowledge, WiDEVIEW is the first dataset to integrate UWB technology for enhanced road users localization in urban driving environment for autonomous cars. 

\section{Sensor setup}
\subsection{Sensor Suite and System Setup}
\begin{figure}
    \centering
        \includegraphics[width=0.7\textwidth,trim=0cm 5cm 0cm 1cm,clip=true]{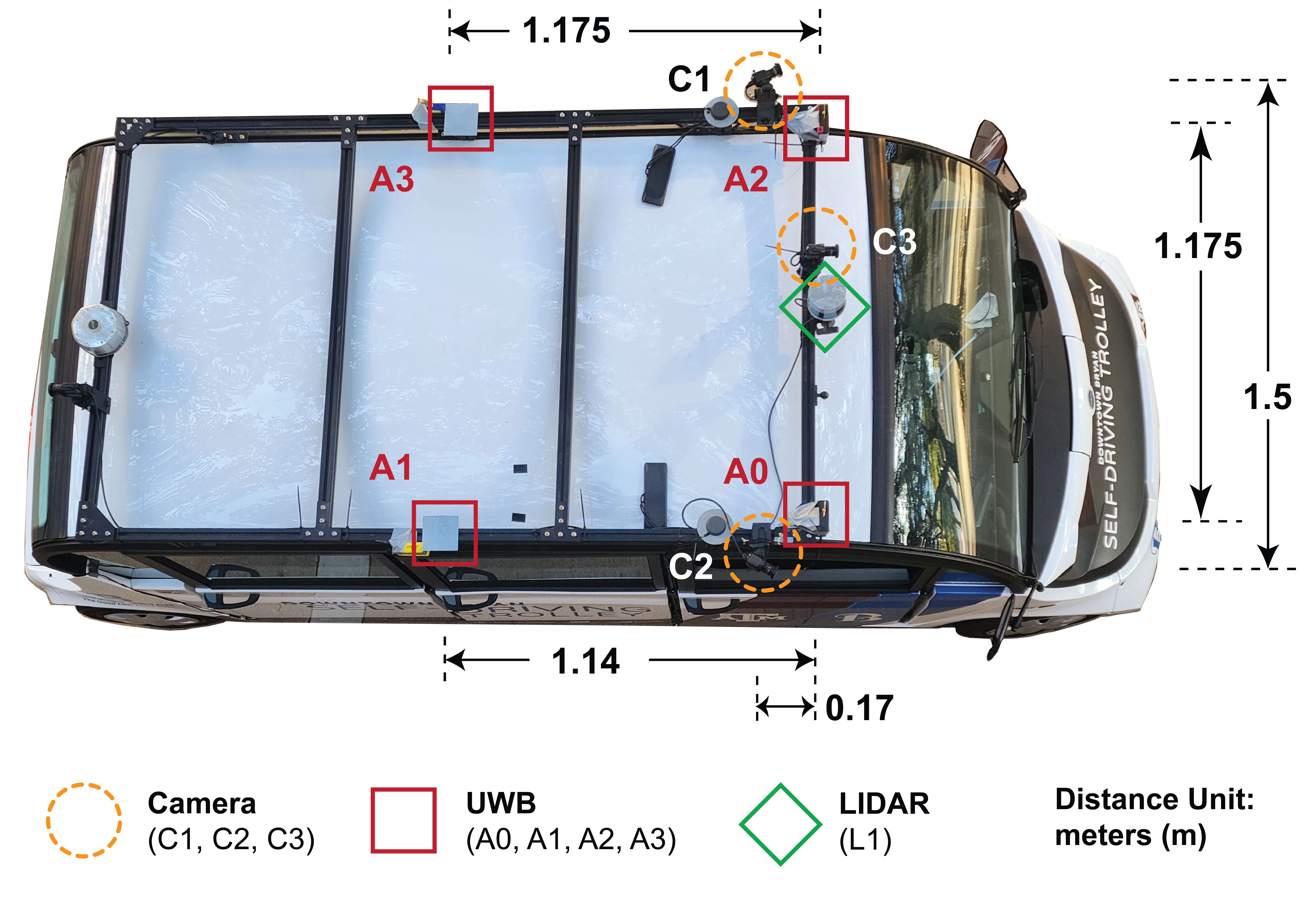}
        \caption{Top view of sensor setup on the self-driving shuttle.}
        \label{fig:golfcart}
\end{figure}

Our sensor setup is illustrated in Fig. \ref{fig:golfcart} and includes:
\begin{itemize}
    \item \textbf{3 X point grey RGB cameras}: 15FPS, 1296x728 resolution.
    \item \textbf{1 X Velodyne Ultra Puck}: 32 Channels, 10hz, 40 $\deg$ vertical field of view.
    \item \textbf{Inertial Navigation System}: Vectornav VN-300 Dual Antenna GNSS/INS, 100Hz INS and 50Hz raw IMU.
    \item \textbf{UWB sensors}: Decawave (now Qorvo) TREK1000 RTLS evaluation kit with DW1000 UWB chip configured as 4 anchors and 6 tags.
\end{itemize}
Fig. \ref{fig:golfcart} denotes the configuration of the sensor suite mounted on Polaris GEM e6. 

   

\subsection{UWB Ranging with ToF Measurements }
UWB, a radio-frequency (RF) technology, utilizes low energy pulse wireless communication typically for short range, high bandwidth applications. By measuring time of flight (ToF) across various frequencies, it is possible to measure distance between modules while overcoming issues typically associated with multipath errors. This has allowed UWB modules to be applied towards localization and tracking problems. 

Decawave modules used in this paper estimate anchor-tag range using the Double-Sided Two Way Ranging (DS-TWR). Reader is encouraged to refer to \cite{uwb_ref} for detailed range calculation equations. 

\subsection{Camera Calibration}
The sensor suite cameras (left, middle and right on the autonomous shuttle roof rack) have a 30$^{\circ}$ overlapping field of view between adjacent cameras. The intrinsic and extrinsic camera parameters are estimated by the multiple camera calibration tool in the Kalibr toolbox \cite{kalibr}. A 6 x 6 aprilgrid target with spacing of 0.03m is used. We utilize a pinhole projection model for our cameras, where a three-dimensional scene is projected onto an image plane through a perspective transform. The calibration details can be found on the dataset Github\footnotemark[1] page.

\section{Dataset}
\begin{table*}[!ht]
\scriptsize
\resizebox{\columnwidth}{!}{
\begin{tabular}{ccccccc} \hline
        %
         \textbf{Scenario} & \textbf{Tag }& \textbf{Pedestrian} & \textbf{Cyclist} & \textbf{Type} & \textbf{Vehicle } & \textbf{Traffic} \\
         \hline
         1 & T0 & crossing, crosswalk & N/A & individual & approaching slowly & light\\
         2 & T0 & crossing diagonally & N/A & individual & approaching fast  &  moderate \\
         3 & T0 & crossing & N/A & individual & turning left &  moderate  \\
         4 & T0 & crossing& N/A & individual & turning right  & moderate  \\
         5 & T0 & crossing NLOS & N/A & individual & approaching & moderate\\
         6 & T1, T2  & crossing consecutively& N/A & individuals & approaching, turning right & light  \\
         7 & T1, T2 & N/A & overtake, crossing, riding & individuals & moving forward &  light  \\
         8 & T1, T2 & N/A & crossing NLOS & group & approaching & moderate  \\
         9 & T0, T1, T2 & crossing, opposite & N/A & individual, group  & stationary & moderate\\
         10 & T0, T1, T2 & crossing, opposite & N/A & individual, group & approaching slowly  & moderate \\
         11 & T0, T1, T2 & crossing, walking & N/A & individuals & approaching, turning left & light \\
        12 & T0, T1, T2 & standing, curb & moving forward & individual, group & following  & light  \\ 
        13 & T0, T1, T2 & crossing diagonally & riding together, then split & individuals, group& approaching fast  & light  \\ 
         14 & T0, T1, T2 & crossing diagonally & crossing, turning right & individuals & approaching slowly& light  \\
         15 & T0, T1, T2 & crossing, crosswalk & crossing, crosswalk & individual, group & stationary & light  \\
         16 & T0,T3,T4,T5 & crossing LOS/NLOS & N/A & groups & approaching   & moderate \\
         17 & T0,T3,T4,T5 & crossing & N/A & group & approaching & moderate \\
         18 & T0,T2,T3,T5 & crossing LOS/NLOS & crossing NLOS & groups & approaching & moderate \\
         19 & T0,T2,T3,T5 & crossing LOS,standing & crossing LOS & groups & approaching & moderate \\
         20 & T0$\sim$T5 & crossing, crosswalk& N/A & groups & stationary  & heavy \\
         21 & T0$\sim$T5 & crossing, crosswalk & N/A & groups & approaching &  heavy \\
         22 & T0$\sim$T5 & crossing, standing & N/A & groups & approaching, U-turn  & heavy \\
         23 & T0$\sim$T5 & crossing & crossing, overtake & individuals, group & approaching, U-turn & heavy \\
        \hline
\end{tabular}}
\caption{Summary of data collection conditions and scenarios.}
\label{tab:scenario_table}
\end{table*}
\subsection{Data Description}
The data was collected on and around a university campus under light, moderate and heavy traffic conditions during daytime with varying illuminations. The traffic conditions (light or heavy) refer to other vehicles on the road. Number of other road users (pedestrians, cyclists, etc.) was significant in all traffic conditions on and around campus. 
The sensor data includes RGB images from 3 calibrated cameras, point cloud data from LiDAR and ranging data from 4 UWB modules mounted to the roof rack of the vehicle. Vehicle information like speed and position was recorded using the vectornav inertial navigation system. The data was recorded in rosbag format. There are 21 bag files in total ranging from 30 to 149 seconds in duration. 

From UWB perspective, data collection focused on two kinds of road users: pedestrians and cyclists. UWB modules mounted on the shuttle were configured as anchors, whereas UWB modules carried by the road users were configured as tags to collect the ranging data with respect to (w.r.t) the 4 anchors. The number of UWB modules configured as tags ranged from a minimum of 1 to maximum 6 across various scenarios to capture both individual and group road users' interaction scenarios with the vehicle. It should be noted that for TREK1000 modules each anchor can communicate with a maximum of 8 tags and each tag can communicate with maximum 4 anchors.

\begin{figure}[!ht]
    \centering
    \begin{subfigure}[b]{0.49\textwidth}
        \centering
        \includegraphics[width=\textwidth]{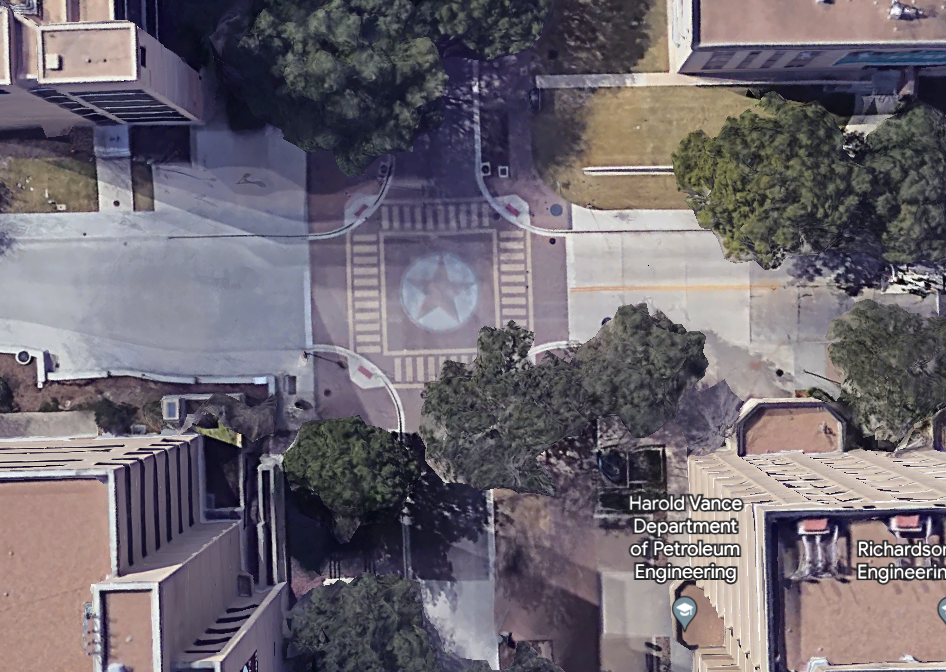}
        \caption{On campus}
        \label{fig:bissel st}
    \end{subfigure}
    \hfill
    \begin{subfigure}[b]{0.49\textwidth}
        \centering
        \includegraphics[width=\textwidth]{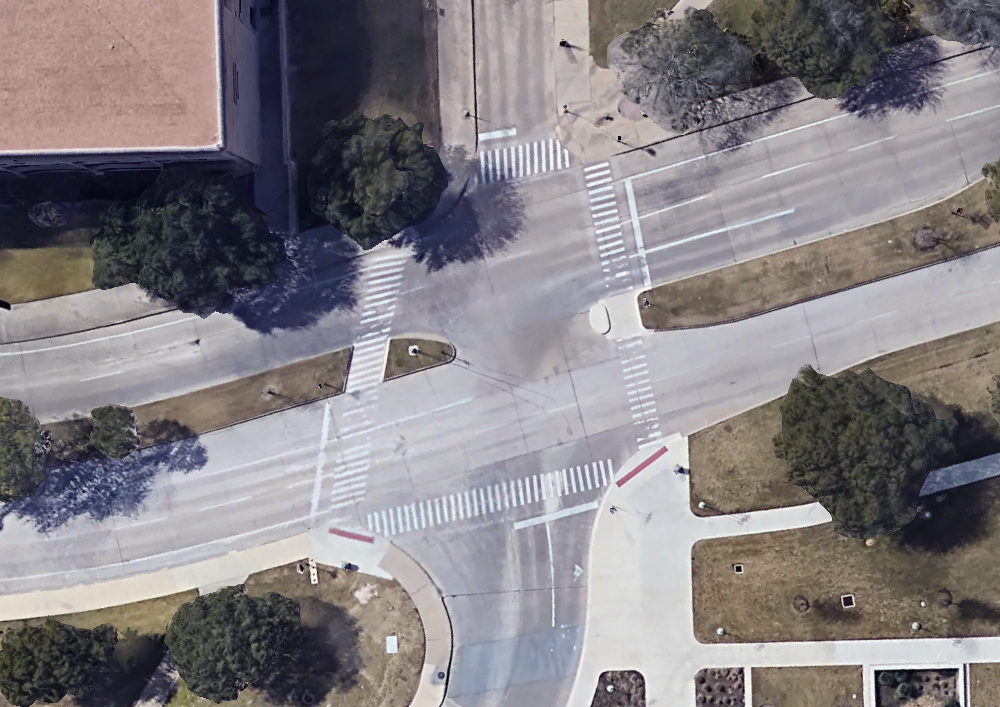}
        \caption{Around campus}
        \label{fig: wisenbaker}
    \end{subfigure} 
        \caption{Data collection scenes examples on and around campus. Both are unsignalized intersections but the latter one is much busier.}
        \label{fig: bissel and web}
\end{figure}

\subsection{Data Annotations}
The dataset contains 30936 frames which come from the front three cameras with 10312 frames each. Every 5-th frame was annotated manually and carefully in Computer Vision Annotation Tool (CVAT\footnote{\url{https://github.com/opencv/cvat}}) for the front middle camera frames while the ones in between are interpolated automatically without adjustment. There are a total of 8368 bounding boxes manually annotated for pedestrians and cyclists along with six tag ID for those who carried tags, and 33239 bounding boxes are interpolated in CVAT. These annotations are exported in MOT 1.1 format for multi-object tracking evaluation. The annotations for the left and right cameras frames are in progress and will be updated on the Github\footnotemark[1] gradually.

\subsection{Data Collection Scenarios}
A summary of data collection scenarios is presented in Table \ref{tab:scenario_table}. For the header of each column, \emph{Tag} denotes the tag ID used in the scenario. \emph{Pedestrian} and \emph{Cyclist} describe the keyword for road users' actions, \emph{N/A} means that particular road user (pedestrian or cyclist) is not a part of that scenario. \emph{Type} shows whether the tags behave as individual or as group. \emph{Vehicle} represents the vehicle actions. \emph{Traffic} reflects the condition of the traffic in light, moderate, and heavy. Data collection scenarios aim at capturing crucial vehicle-pedestrian, and vehicle-cyclist interactions from a safety perspective. Further, the scenarios are chosen to evaluate the robustness of UWB technology and its comparison with vision depending upon a number of factors such as the speed of the vehicle, traffic density, line-of-sight (LOS), and non-line-of-sight (NLOS) conditions (light and heavy obstructions). In addition, varying number of tags are used in different scenarios to evaluate the anchor-tag communication performance. Heavy/Moderate traffic scenarios are challenging as there are more vehicles on the road, which adds obstructions to the existing environment. In light traffic scenarios, the vehicle has a higher speed relatively. The maximum recorded vehicle speed for high-speed scenarios is 10 m/s and approximately 6 m/s for low-speed scenarios. It should be noted that since most of the data focuses on road users' localization, the vehicle speed was reduced significantly (e.g. 2 m/s) in proximity to pedestrians and cyclists at the crossing. The scenarios which mention the vehicle as stationary denote the cases where the vehicle stops and waits for road users to cross the road. Overall, these scenarios are useful to evaluate UWB performance in a combination of vehicle/road user trajectories at different vehicle speeds and traffic and environment conditions. Fig. \ref{fig: bissel and web} shows the google earth screenshots of the two locations used for data collection. As given in Table \ref{tab:scenario_table}, certain scenarios are unique whereas some of the others are similar to each other in terms of road users and vehicle actions. Multiple sequences of similar scenarios are collected to test and ensure reliability and consistency of UWB performance.

\section{Methodology}
\subsection{2D Multi-Object Tracking}
The dataset is evaluated on YOLOv5 [24] and DeepSORT [4] algorithms to detect and track multiple pedestrians in video sequences of the front middle camera. The object class to be detected is set to 0 (person) when running inferences based on the weight $crowdhumanyolov5m$ pre-trained on the MS COCO dataset for YOLOv5 and $osnet x1 0$ for DeepSORT.

\subsection{UWB Based Pedestrian/Cyclist Localization}
Once anchor-tag ranges are obtained by each tag, the 2D location of the tag (pedestrian/cyclist) can be calculated by using the Trilateration algorithm.

The mid-point between A0 and A2 in Fig. \ref{fig:golfcart} is considered as the origin for measured distances. A minimum number of anchor-tag ranges required for Trilateration is $3$, for cases where all four anchor-to-tag ranges are available Least Square Estimations (LSE) is used to calculate the optimal tag positions. 

Tag position calculated using the Trilateration algorithm is w.r.t the designated origin point on the vehicle. Since both the vehicle and tags are moving, a common reference origin point is needed to visualize the vehicle and pedestrian trajectories together. This reference point is chosen as the vehicle position at the first instance when the anchor-tag range message is received. This position is obtained from the vectornav INS system in the form of latitude longitude and is converted to UTM coordinates (meters) using the \textrm{utm} python library. Further, in order to calculate the tag position in Northing Easting Direction (NED) coordinates, a rotation matrix is applied to the calculated tag position for transformation, and the current vehicle position offset is added to the calculated tag position. 

The tag localization depends on at least $3$ anchor to tag ranges. However, these ranges are not always available and are prone to data loss, especially in the presence of obstructions or because the vehicle is moving too fast. To get a more accurate estimate of tag position, we use the Kalman Filter (KF) along with Trilateration algorithm for pedestrian 2D position estimation. KF has two steps which are prediction and measurement update. The prediction step estimates the next tag position based on the previous position, and this prediction is further updated whenever the tag position measurement is received by Trilateration. In cases when no measurement is received, the KF continues to predict the tag position for a threshold of $6$ seconds. In case no measurement is received after $6$ seconds, the KF stops estimating till the next measurement is received to avoid error accumulation. It is worth mentioning that state-of-the-art trilateration algorithm and KF have been used for localization since the scope of this paper is to evaluate UWB as a sensing modality for vehicle-pedestrian interactions. Future work will include improvements to the localization algorithms.

\section{Results and Discussion}
\subsection{Object Tracking Results}
\begin{table*}[!h]
\noindent
\scriptsize
\centering
\begin{tabular}{cccccccccccccc}  
        %
         Case & HOTA & MOTA & MOTP & IDF1 &FP &FN & ID Sw. & Recall & Precision & Dets & GT Dets & IDs & GT IDs \\
         \hline
         1 & 41.577 & 42.046 &72.496  & 54.958 & 9144 &11006  & 729  &69.451 & 73.236 &34165 & 36027 & 491 &231  \\
         \hline
         2 & 45.873 & 44.204 &78.173 & 57.293 & 1607 & 2038 & 398 & 71.874  & 76.42 & 6815& 7246 &425& 231  \\
        \hline
\end{tabular}
\caption{\small {2D Multi-Object Tracking results on our dataset by using YOLOv5 and DeepSORT as detector and tracker. Case 1 represents evaluation results when all frames are considered, while in case 2 every 5-th frames are used.}}
\label{tab:MOT_metrics}
\end{table*}
 To evaluate the object tracking performance on our dataset, we report the offical MOT evaluation metrics from \cite{trackeval} which include Higher Order Tracking Accuracy (HOTA), Multiple Object Tracking Accuracy (MOTA), Multiple Object Tracking Precision (MOTP), Identity F1 Score (IDF1), False Positive Rate (FP), False Negative Rate (FN), ID switches (ID Sw.), Recall and Precision, detection counts(Dets), ground truth detection counts(GT Dets), pedestrian ID counts(IDs), ground truth pedestrian ID counts(GT IDs), see Table \ref{tab:MOT_metrics}. Examples of tracking frames can be referred to in Fig. \ref{fig:seq17_vision}. It is worth noting that there are other trackers which might perform better than the one we used, but this is beyond the scope of this paper. As can be seen from the results, the ID switches are a big problem to realize accurate and reliable pedestrian tracking.

\subsection{Range Accuray: LiDAR vs. UWB}
We use LiDAR range data instead of RTK GPS to evaluate the accuracy of UWB range data because of close proximity to obstructions (buildings etc.) in a significant number of scenarios, where the RTK signal will be of poor quality. Moreover, we are concerned with relative accuracy between the vehicle and the pedestrian/cyclist for which LiDAR provides far better accuracy than RTK. 

\begin{figure}[!h]
    \centering
    \begin{subfigure}{0.45\textwidth}
    \includegraphics[width=\textwidth]{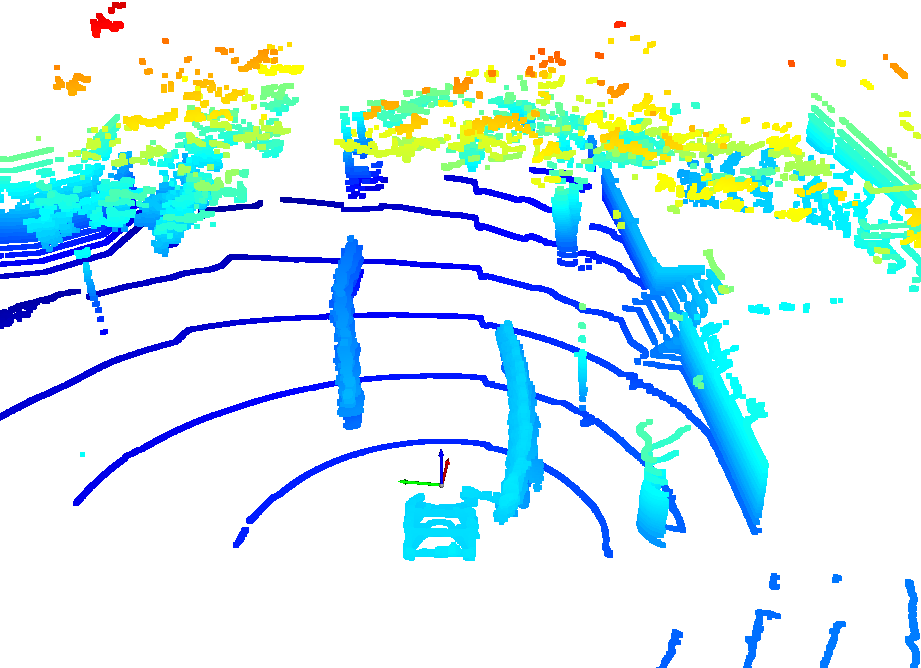}
    \caption{LiDAR point clouds}
    \label{fig: lidar open 3d}
    \end{subfigure}
    \begin{subfigure}{0.45\textwidth}
    \includegraphics[width=\textwidth]{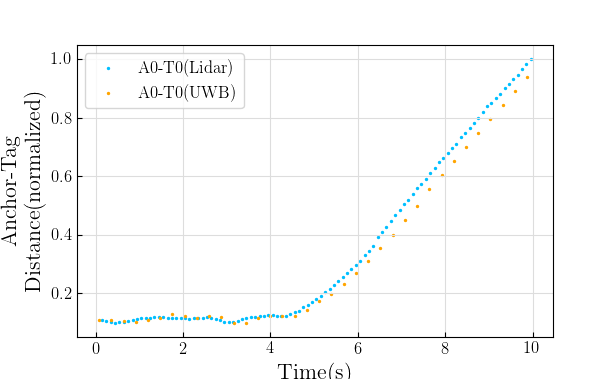}
    \caption{A0-T0 distances vs. Time}
    \label{fig: range_lidar_uwb}
    \end{subfigure}
    \caption{Comparison of anchor 0 to tag 0 (A0-T0) distances that are obtained from UWB range data and LiDAR point cloud data.}
    \label{fig:lidar vs uwb}
\end{figure}

For comparison between LiDAR and UWB accuracy we present the results for one of the sequences in LOS conditions.
Fig. \ref{fig: lidar open 3d} shows the LiDAR point clouds visualized in Open3d, which depicts a scenario when two pedestrians are walking on two sides of the parked vehicle. The pedestrian on the right is tag 0 (T0). A list of T0 locations from LiDAR point clouds which comes from the first 10 seconds of the first data sequence, can be expressed w.r.t anchor 0 (A0) as, 
\begin{align}
    (x_{A0T0},y_{A0T0},z_{A0T0}) &= (x,y+\frac{1.175}{2},z)
\end{align}
where $(x,y,z)$ is the pedestrian location w.r.t LiDAR mounted on the vehicle and $\frac{1.175}{2}$ is the absolute relative distance between LiDAR and A0.
Then the A0 and T0 distance can be calculated by:
\begin{align}
    Distance_{A0T0} &= \sqrt{{x_{A0T0}}^2+{y_{A0T0}}^2+{z_{A0T0}}^2}
\end{align}
 Fig. \ref{fig: range_lidar_uwb} shows the comparison of A0 to T0 distances that are obtained from UWB range data and LiDAR point cloud data, respectively. \textbf{The average magnitude of distance error between LiDAR and UWB is found out to be 0.1983 meters (m) with a standard deviation of 0.1996 m.}

\subsection{UWB Range and Localization Results}\label{uwb_results}
In this section, UWB ranges and localization results are discussed for typical scenarios of the dataset, where the UWB technology can be useful as a complement of vision for safer navigation. Current shortcomings and the scope of improvement are also discussed. 
Normalized range and trajectory plots are presented for the discussed scenarios. For each scenario, normalization is done w.r.t the maximum range observed for that scenario for range plots, and for localization plots w.r.t the maximum northing/easting coordinate observed in reference to the vehicle starting point. Normalization is useful as we are only concerned with relative localization between vehicle and pedestrians. Further, the maximum range reliable for UWB is subjective to the sensor used and its specifications. Thus, the discussion focuses on a qualitative comparison of vision and UWB and the scenarios where UWB addition can provide improved information.
\begin{figure}[!h]
    \centering
    \begin{subfigure}{0.85\textwidth}
    \includegraphics[width=\textwidth]{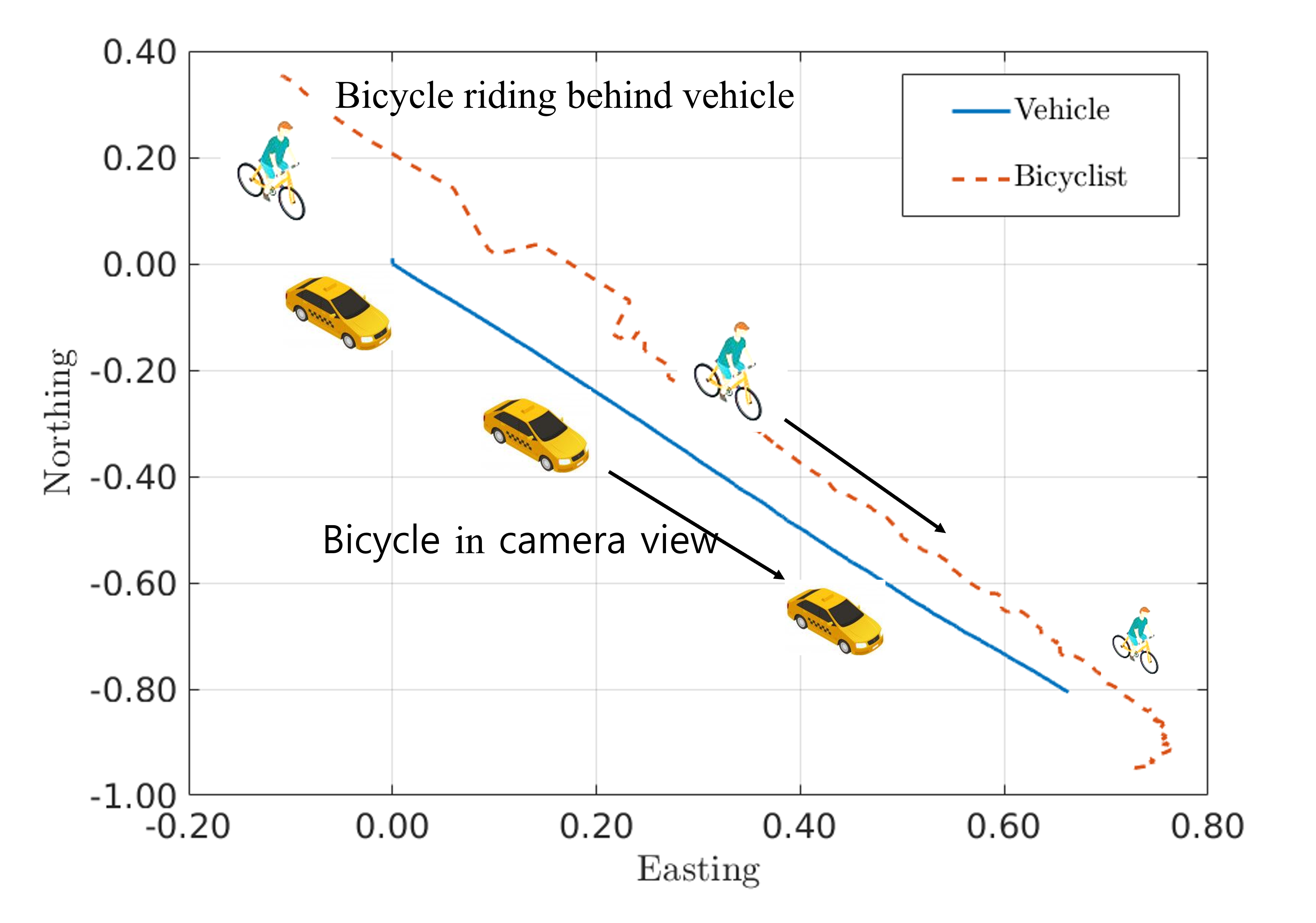}
        \caption{Cyclist and vehicle trajectories}
        \label{fig:loc_seq8}
   \end{subfigure}
    \begin{subfigure}{0.85\textwidth}
    \includegraphics[width=\textwidth]{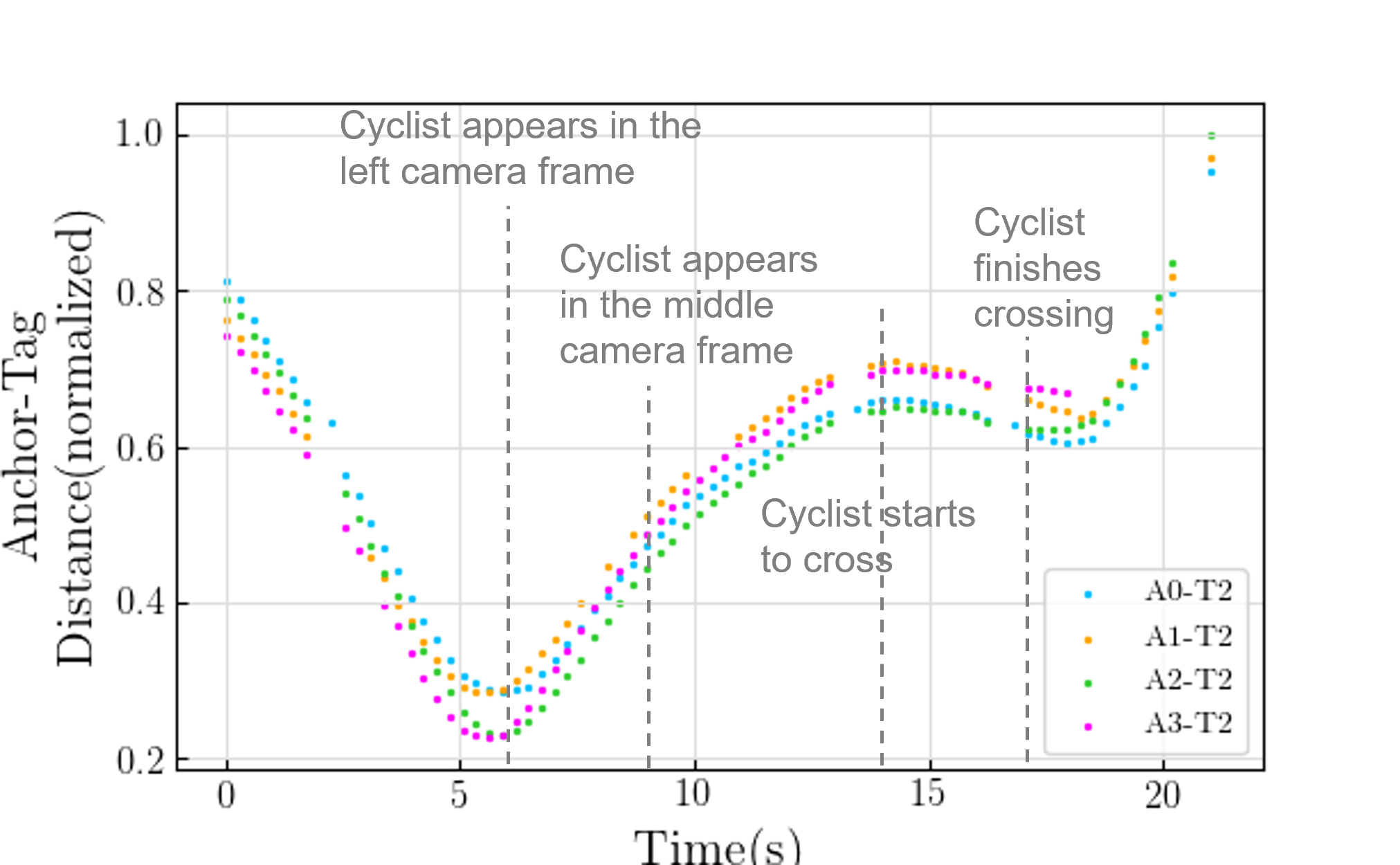}
        \caption{Anchor to tag ranges for T2 (cyclist) }
        \label{fig:range_seq8}
    \end{subfigure}
    \caption{Scenario 7: Tag 2 (T2) localization and range results. }
        \label{fig: scenario7}
\end{figure}

Fig. \ref{fig:loc_seq8} presents the trajectory of a cyclist/Tag 2 (T2) and that of the vehicle. The cyclist starts from behind the vehicle, overtakes the vehicle, and further crosses in front of the vehicle to the right while the vehicle is moving forward. However, the cyclist is not detected in the front middle camera until he overtakes the vehicle in the second half of the scenario. Meanwhile, UWB is consistently providing range data ( and thus tag position) even when a cyclist is behind the vehicle, see Fig. \ref{fig:range_seq8}. This scenario is quite common when the autonomous vehicle is equipped with only one camera thus having limited field of view or blind spots. With the help of UWB sensors, the vehicle can be aware of the approaching road users even when they have not come into view yet, thus executing timely control commands.

\begin{figure}[!h]
    \centering
    \begin{subfigure}{0.85\textwidth}
    \includegraphics[width=\textwidth]{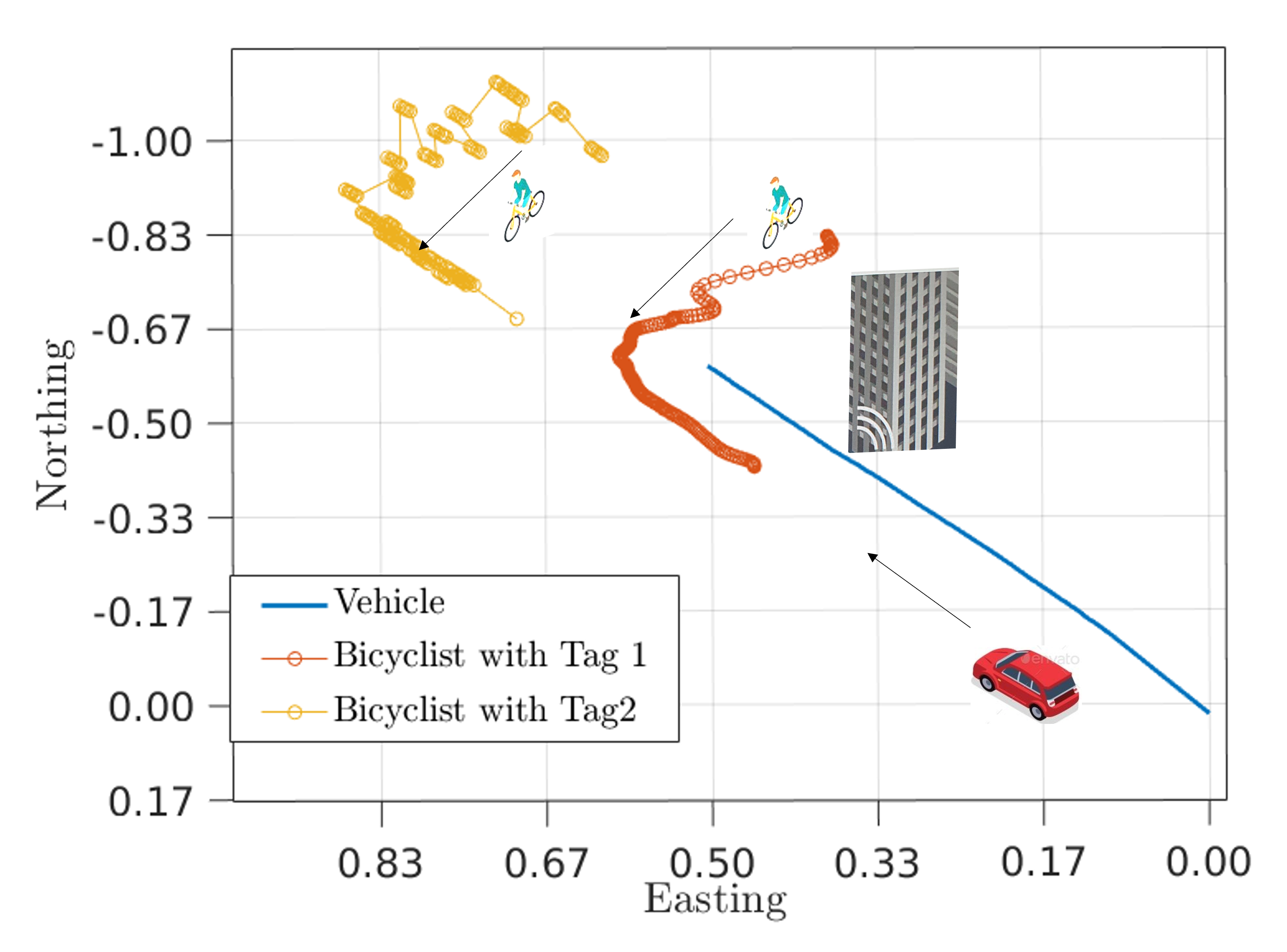}
        \caption{Scenario 8: complete obstruction }
        \label{fig:seq18_NLOS_cyclist}
    \end{subfigure}
    \begin{subfigure}{0.85\textwidth}
    \includegraphics[width=\textwidth]{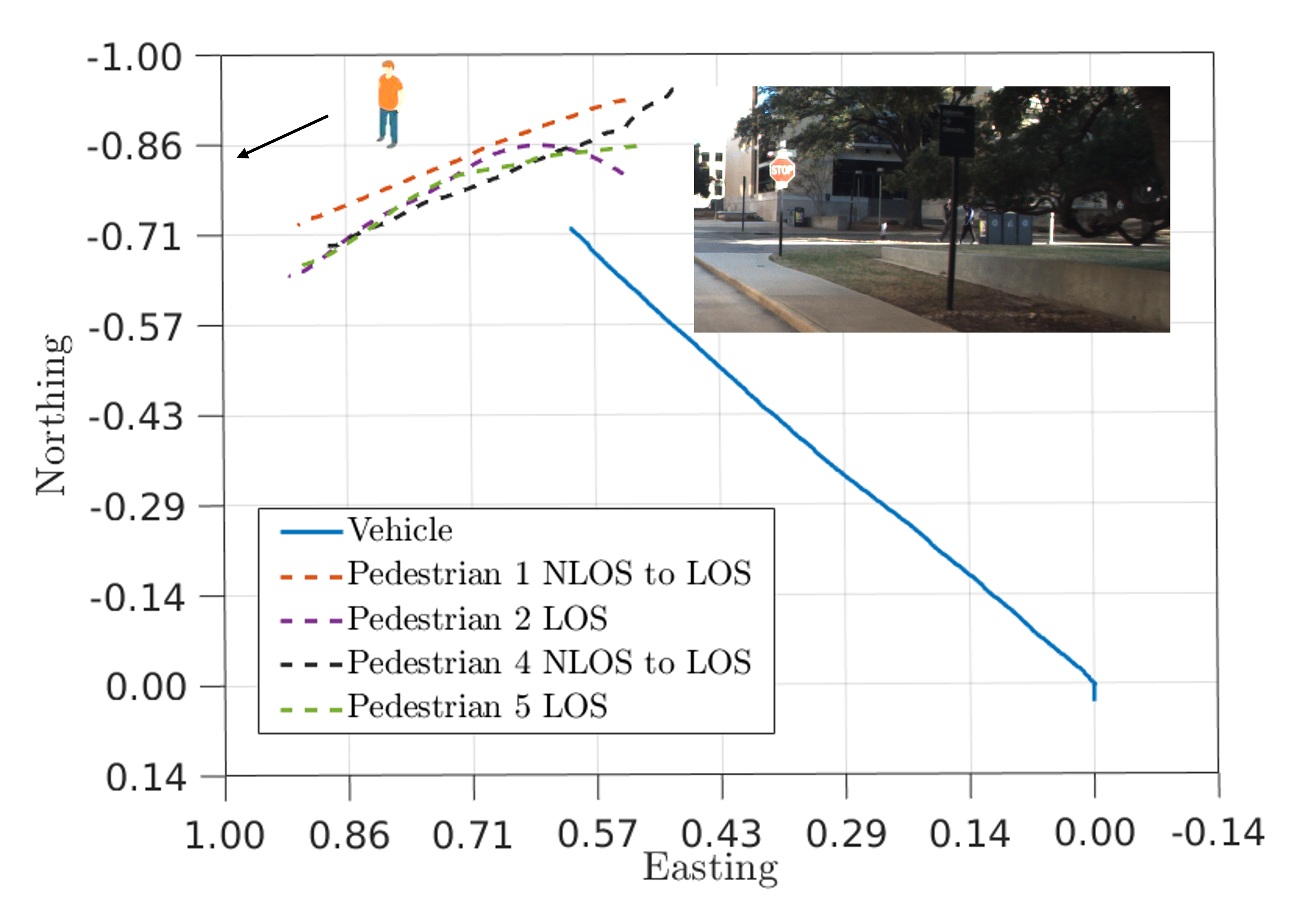}
        \caption{Scenario 16: LOS vs. NLOS with light obstruction}
        \label{fig:seq18_NLOS_LOS_ped}
    \end{subfigure}
    \caption{NLOS scenarios with obstructions.}
        \label{fig: NlOS}
\end{figure}

We also evaluate the position estimation in NLOS conditions with complete obstruction, which is shown in Fig. \ref{fig:seq18_NLOS_cyclist}. In this scenario, two cyclists are behind the building on the right at the beginning and come out of the obstruction into LOS to cross in front of the vehicle by taking a left turn. Tag 2 position estimation is worse than Tag 1 because Tag 2 is further behind the building and Tag 1 is closer to the road and appears in LOS sooner. This scenario depicts a very critical application of UWB technology especially in a campus/urban driving scenario since the vehicle will not have any information about the obstructed cyclist and lack of this information can result in a collision accident. Even though the data drops are relatively higher for this case, both cyclists do get sensed by the UWB $3$ seconds in advance despite the obstruction than getting detected by the camera.

Fig. \ref{fig:seq18_NLOS_LOS_ped} includes a scenario to evaluate and compare the performance of pedestrian position estimation in LOS versus NLOS with light obstruction. It also aims to evaluate the anchor-tag communication performance when more number of tags are involved. Two tags are barely visible to the vehicle camera and are crossing from behind the trees (light obstruction), whereas the remaining two tags are in LOS. The number of data drops observed for this scenario is just 5\% of the drops observed in scenarios in Fig. \ref{fig:seq18_NLOS_cyclist}, which can be explained by a combination of the nature of the obstruction as well as the difference in moving speed of the tags in two scenarios. Higher speed of cyclists along with heavier obstruction in Fig. \ref{fig:seq18_NLOS_cyclist} results in more range data drops thus worse position estimation compared to the scenario in Fig. \ref{fig:seq18_NLOS_LOS_ped}. 

\begin{figure}[!h]
\centering
\begin{subfigure}{0.45\textwidth} 
\includegraphics[width=\textwidth]{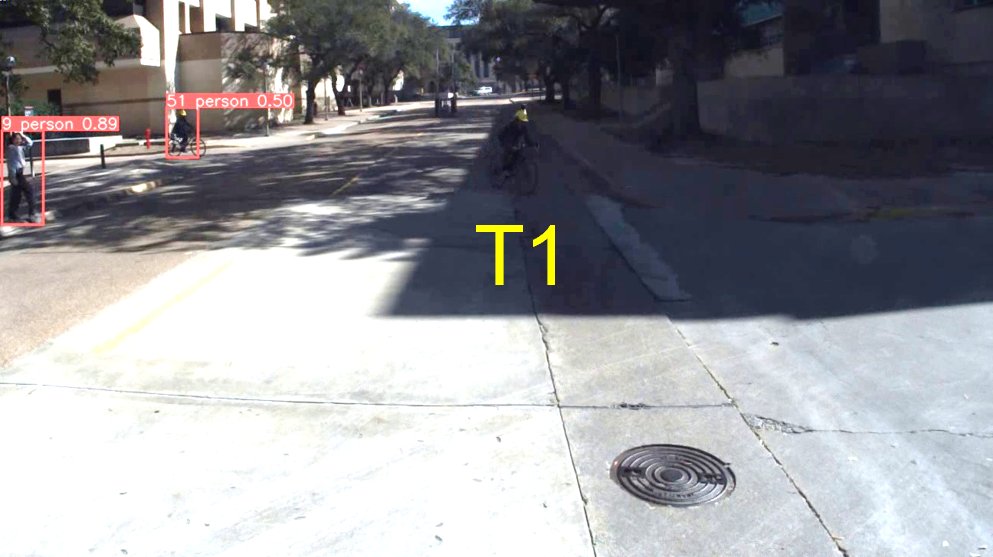} 
\caption{T1 not detected in shades}
\label{fig: not detected_vision}
\end{subfigure}
\begin{subfigure}{0.70\textwidth}
\includegraphics[width=\textwidth]{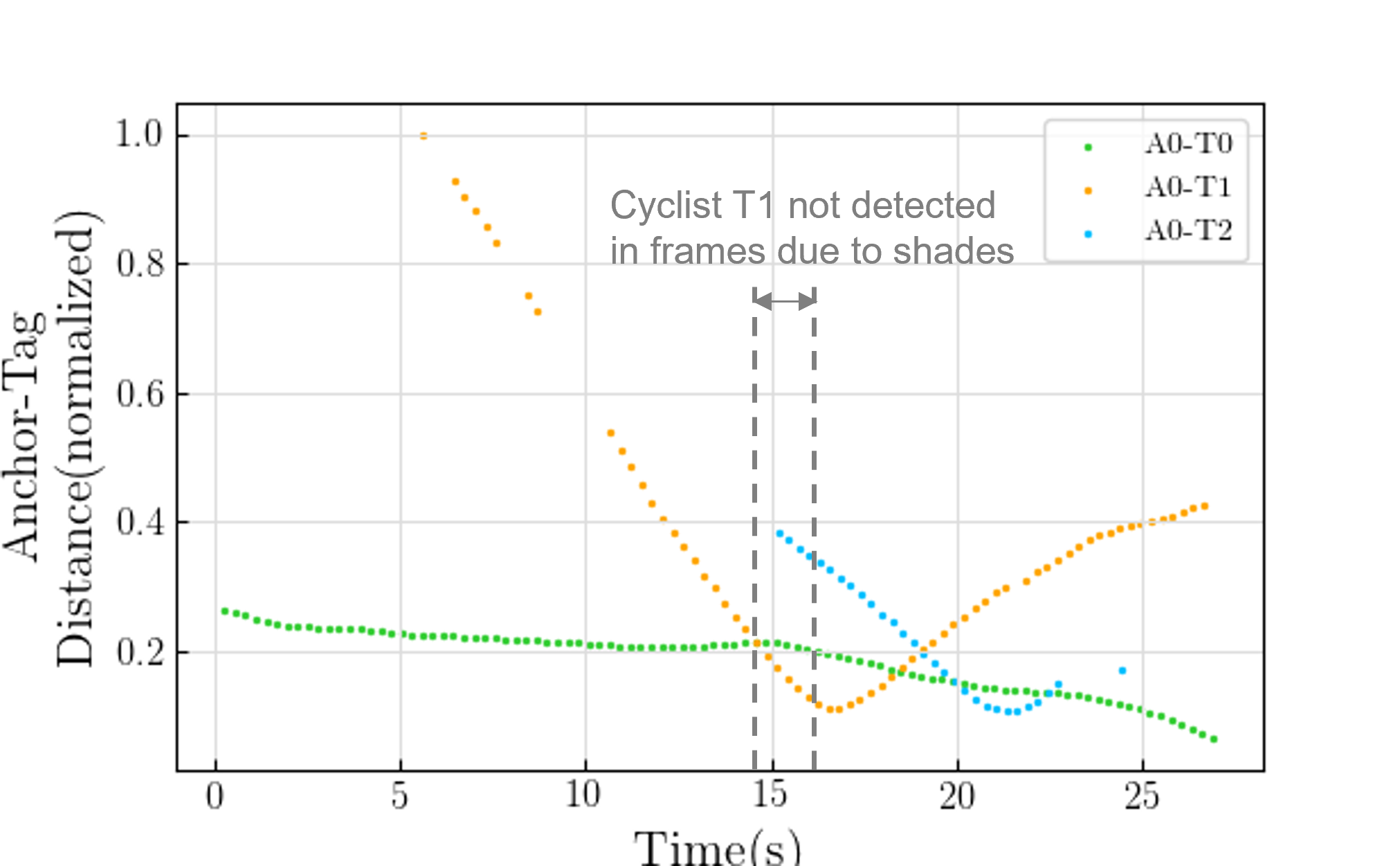} 
\caption{Anchor to tag range results}
\label{fig: not detected_range}
\end{subfigure}
\caption{Scenario 14: vision and range results under bad illumination condition.}
\label{fig: not detected}
\end{figure}
Another scenario in Fig.\ref{fig: not detected} shows an advantage of UWB sensor over RGB camera under bad illumination conditions. Cyclist with Tag 1 (T1) is coming towards the moving vehicle and crossing to the right branch street in front of the vehicle. However, he is not detected at all for almost 2 seconds in the front shades, which can be a dangerous situation since the vehicle will just move forward without timely information. Hence, the UWB data provided during that period will remedy the vision failures.

\begin{figure*}[!h]
\begin{subfigure}{\textwidth}
\includegraphics[width=0.24\textwidth]{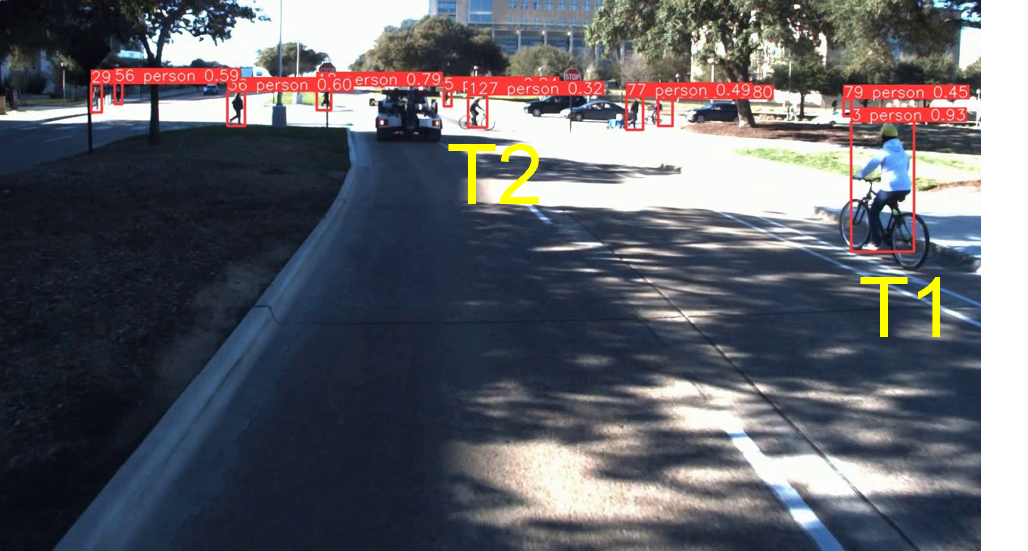} 
\includegraphics[width=0.24\textwidth]{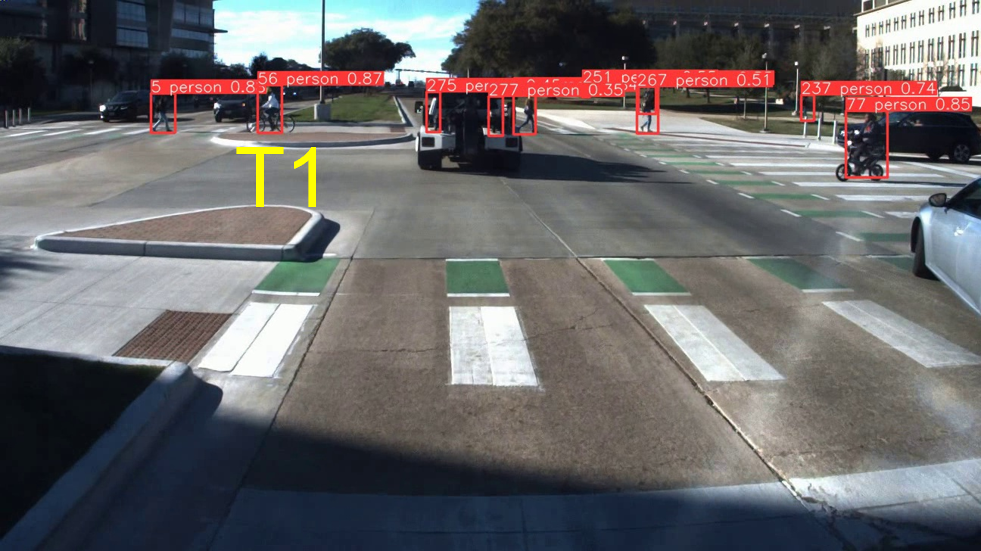} 
\includegraphics[width=0.24\textwidth]{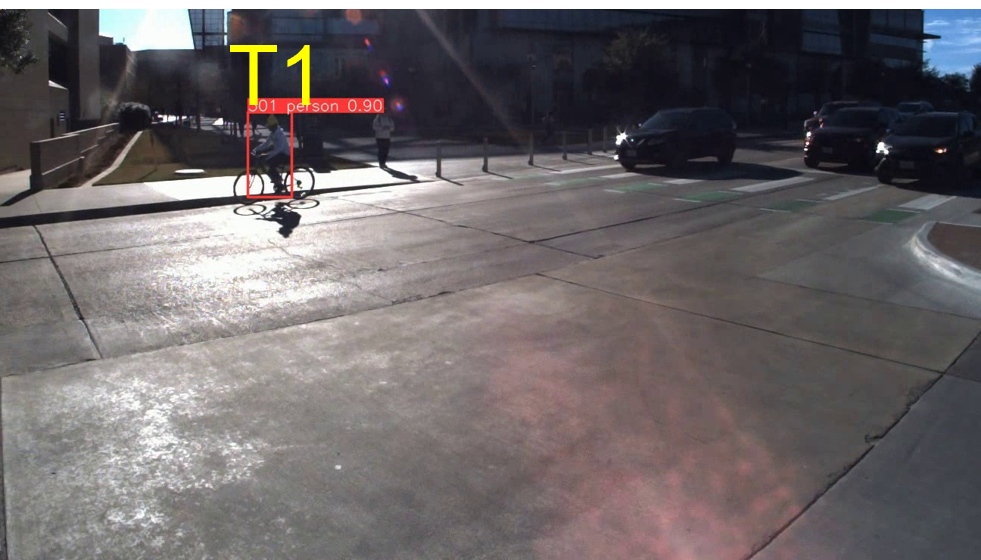}
\includegraphics[width=0.24\textwidth]{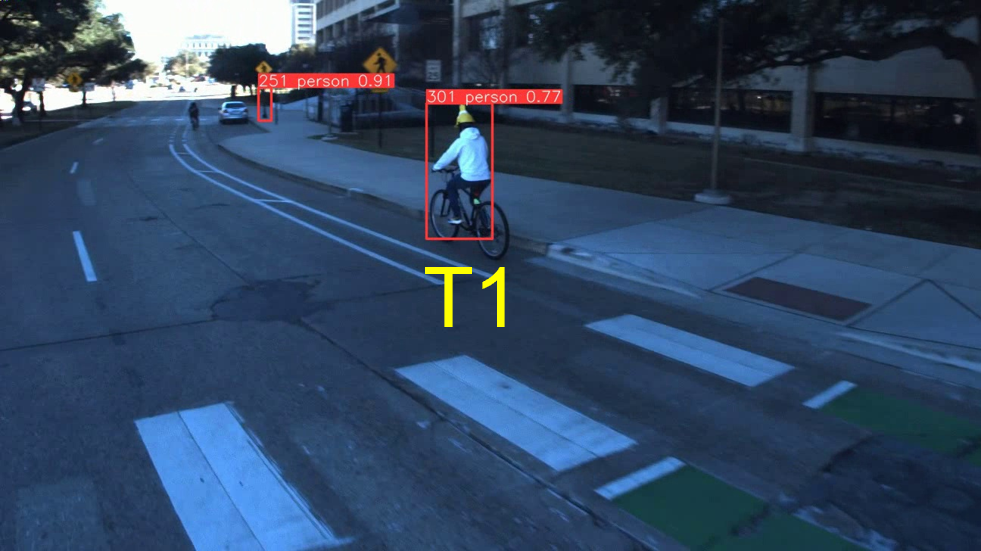} 
\caption{Tracking results in the key middle camera frames}
\label{fig:seq17_vision}
\end{subfigure}
\begin{subfigure}{0.32\textwidth}
\includegraphics[width=\textwidth]{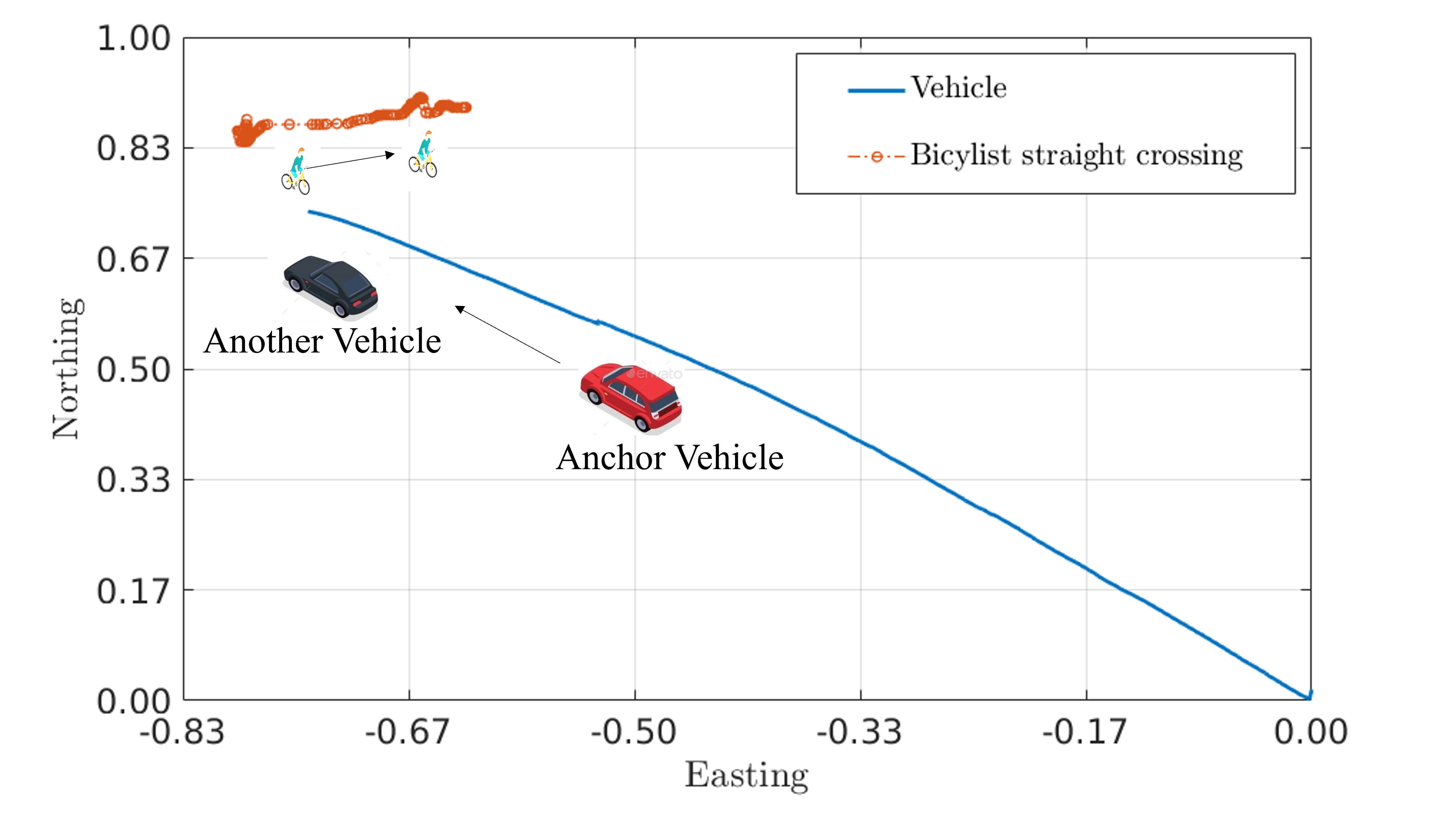} 
\caption{ A cyclist (T2) is crossing while the vehicle is moving forward}
\label{fig:seq17_cyc_crossing}
\end{subfigure}
\hfill
\begin{subfigure}{0.32\textwidth}
\includegraphics[width=\textwidth]{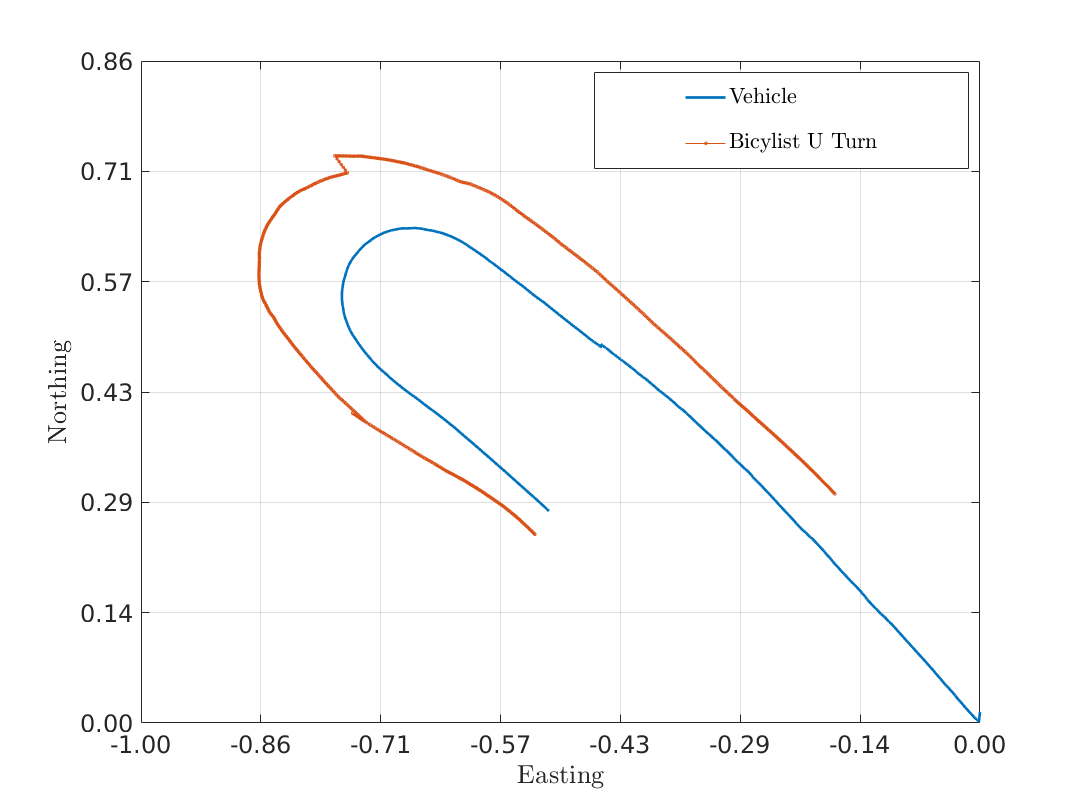}
\caption{ A cyclist (T1) is taking a U-turn while the vehicle is also taking a U-turn}
\label{fig:seq17_cyc_u-turn}
\end{subfigure}
\hfill
\begin{subfigure}{0.32\textwidth}
\includegraphics[width=\textwidth]{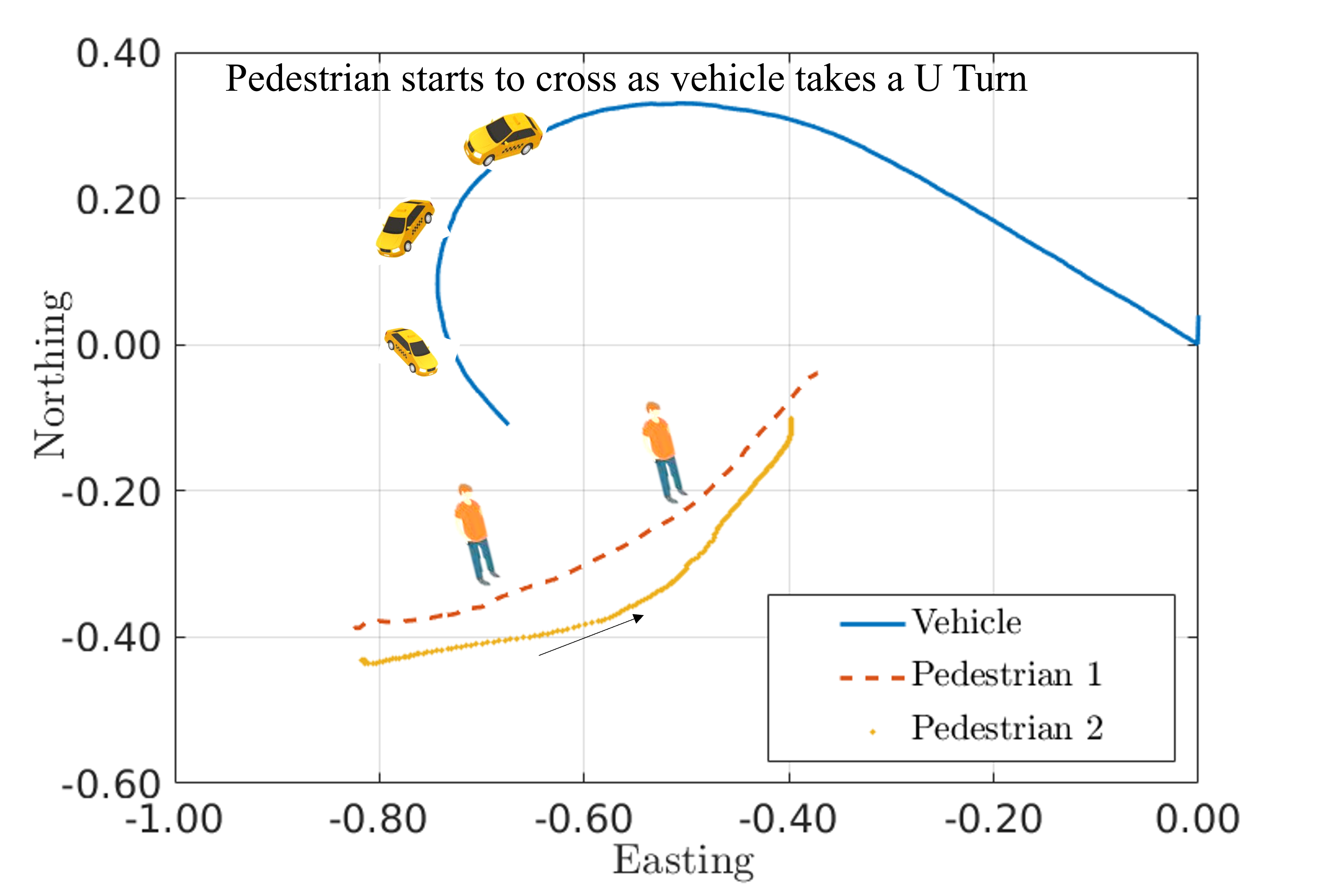} 
\caption{ Pedestrians (T0 and T3) start to cross while the vehicle is taking a U-turn}
\label{fig:seq17_ped_crossing}
\end{subfigure}
\begin{subfigure}{0.45\textwidth}
\includegraphics[width= \textwidth]{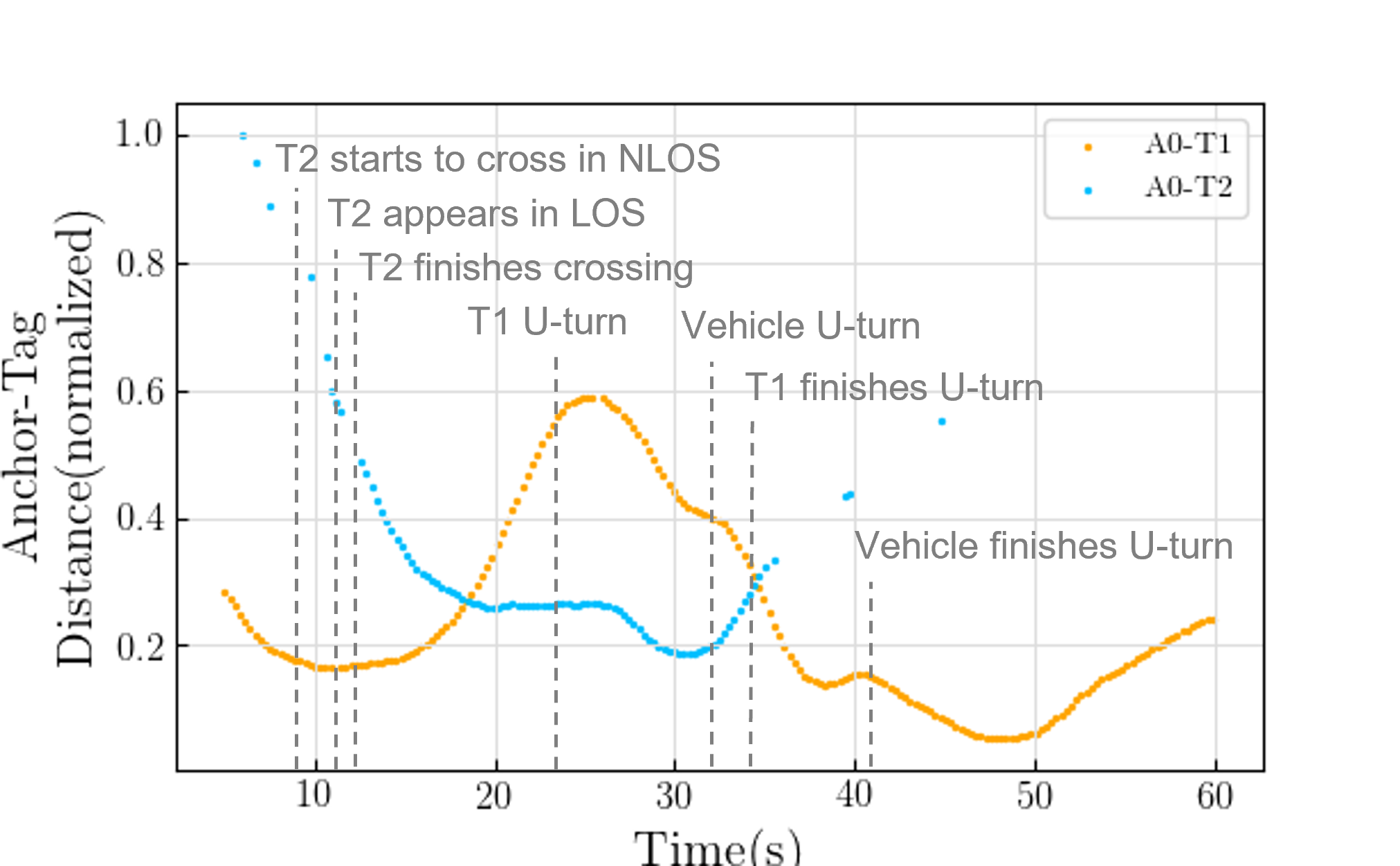}
\caption{Anchor 0 to Tag range results for two cyclists}
\label{fig:seq17_cyc_range}
\end{subfigure}
\begin{subfigure}{0.45\textwidth}
\includegraphics[width=\textwidth]{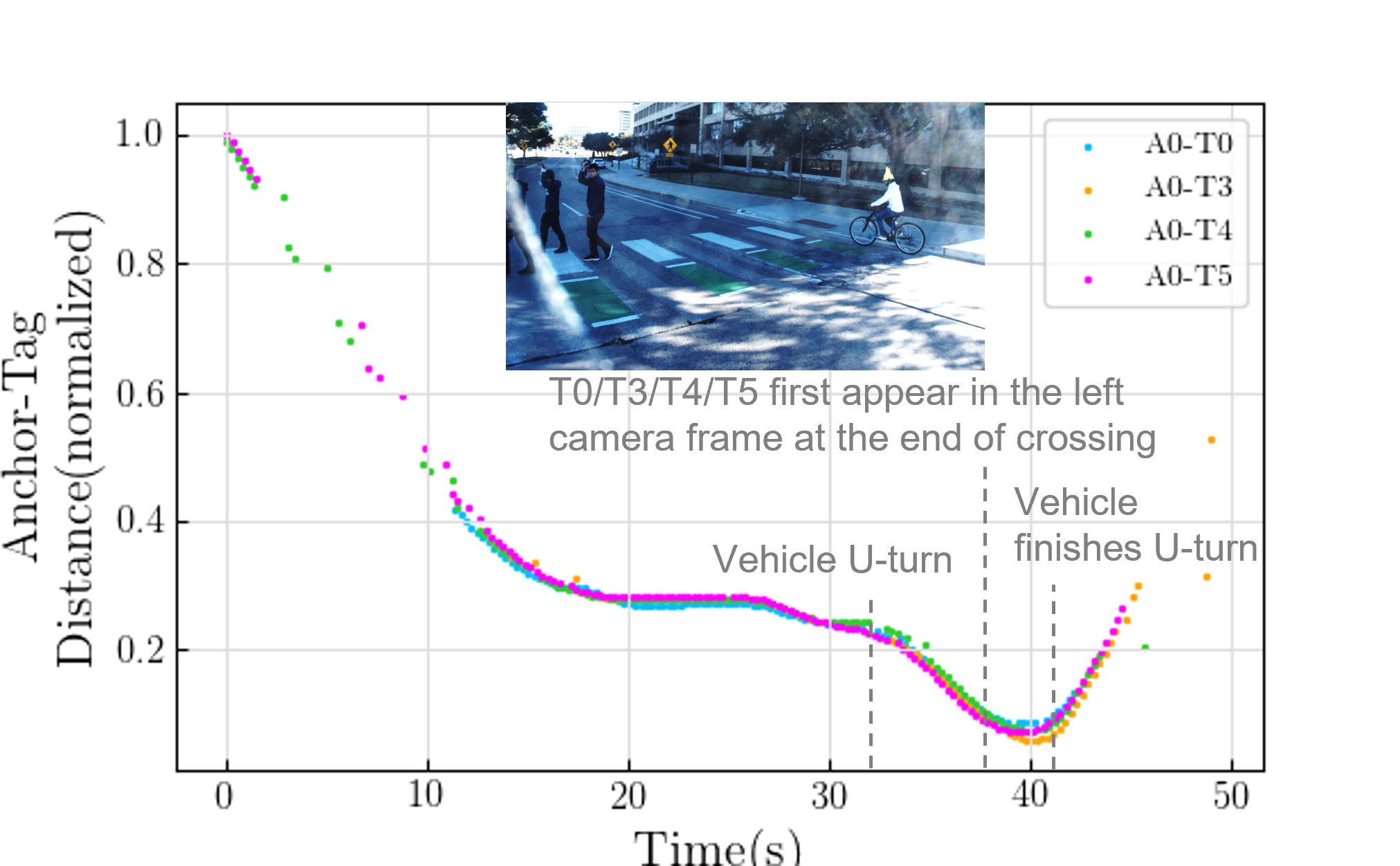}
\caption{Anchor 0 to Tag range results for four pedestrians }
\label{fig:seq17_ped_range}
\end{subfigure}
\caption{Scenario 23: vision, localization, and range results.}
\label{fig: scenario_23}
\end{figure*}

Fig. \ref{fig: scenario_23} shows another scenario recorded off campus in relatively busier traffic conditions, where the vehicle is moving towards an intersection and then making a U-turn to the left side of the road. Fig. \ref{fig:seq17_vision} presents the object tracking results for the key frames of the front middle camera. 
Fig. \ref{fig:seq17_cyc_crossing} represents the trajectory of a cyclist (T2) who is crossing in front of the vehicle from left to right. As it can be seen from Fig. \ref{fig:seq17_cyc_range}, there is a relatively higher range data drop at the beginning of the crossing. This is due to cyclist LOS being completely obstructed by another car directly in front of the self-driving shuttle. Even with a relatively higher frequency of anchor-tag range loss, the cyclist's position is estimated reasonably well. 
Another cyclist (T1) and the vehicle trajectories are shown in Fig. \ref{fig:seq17_cyc_u-turn}, where the cyclist (T1) moves parallel to the vehicle and starts taking a U turn slightly before the vehicle.
Fig. \ref{fig:seq17_ped_crossing} shows the vehicle and pedestrians trajectories obtained from the localization algorithm, where the pedestrians are crossing an intersection as vehicle is taking a U turn. It should be noted that the pedestrians do not appear in the middle camera frame throughout the time when the vehicle is taking a U turn, and even the left camera barely manages to capture pedestrians. However, UWB provides continuous range data in Fig. \ref{fig:seq17_ped_range} and thus timely location of the pedestrians which can be useful to estimate collision probability and safe navigation action of the vehicle. It should be noted that in this particular case, the vehicle did not stop or significantly reduce the speed while approaching the crossing.

\subsection{Discussion}
To summarize, UWB is capable of adding more robustness to the currently existing perception system for vehicle-pedestrian interaction for autonomous driving. For NLOS conditions, we observed that UWB was able to provide information 2 to 3 seconds in advance for light obstructions and 1.5 to 2 seconds in advance for heavy obstructions compared to vision. For direct LOS conditions (when the road user is in camera view), although UWB didn't provide a significant advantage in terms of earlier sensing compared to vision, redundancy in sensing systems contributes towards making the whole system more robust. Further in LOS conditions, intermittent detection/tracking failures can often occur due to poor illumination conditions, camera blind spots in case of moving as well as parked vehicles, UWB sensing provides a very strong use case for safer vehicle-pedestrian interactions by providing timely range and position data up to $40$ meters from the vehicle. 
Furthermore, pedestrians passing each other or walking side by side as a group pose a big challenge for standard object tracking algorithms in the form of occlusions and tracking ID switches, thus resulting in inconsistent and false pedestrian tracking, which will further make the pedestrian behavior or action prediction inaccurate. Given the UWB range and location data are unique to each tag ID, it can be combined with vision data to provide more accurate pedestrian trajectories and potentially avoid the tracking ID losses/switches issues.
Another interesting use case of UWB technology  for vehicle-vehicle-pedestrian interactions is where anchor vehicle in LOS with road users can provide ahead-of-time information to another approaching vehicle which is not in LOS with the road user tag by UWB communication. Though challenging, this provides a strong application for UWB sensing-based collaborative localization for safer navigation and vehicle(s)-pedestrian interactions.

\section{Conclusion}
We introduce WiDEVIEW, a comprehensive multi-modality dataset consisting LiDAR, three RGB cameras, GPS/IMU and Ultra-wideband (UWB) sensor with a focus on vehicle-pedestrian interaction in campus/urban autonomous driving scenarios. The dataset was evaluated on standard vision-based object detection and tracking algorithms. UWB range accuracy was evaluated with LiDAR. UWB based position estimation of road users using state of the art trilateration algorithm with KF estimation was evaluated for both LOS and NLOS scenarios. Future work will include improvements to the localization algorithm and a thorough evaluation of localization accuracy. 
When compared with vision, although UWB as a standalone sensing modality doesn't offer the visualization intuitiveness of a camera, its augmentation can benefit vision systems in scenarios of limited FOV, blind spots, obstructions, adverse lighting conditions, etc. 
Interesting use cases and future extensions of this dataset include investigating the fusion of UWB data with vision to provide robust and reliable tracking of road users and UWB application for vehicle-vehicle-pedestrian interaction scenarios. 





\bibliographystyle{IEEEtran}
\bibliography{References}

\end{document}